\pdfoutput=1

\documentclass[11pt]{article}

\usepackage[]{EMNLP2022}

\usepackage{times}
\usepackage{latexsym}
\usepackage[T1]{fontenc}
\usepackage[utf8]{inputenc}
\usepackage{microtype}
\usepackage{inconsolata}

\usepackage{amsmath}
\usepackage{adjustbox}
\usepackage{enumitem}
\setlist[enumerate]{itemsep=-1mm}

\title{Hyperdecoders: Instance-specific decoders for multi-task NLP}

\author{Hamish Ivison \and Matthew E. Peters\\
  Allen Institute for AI \\
  \texttt{\{hamishi, matthewp\}@allenai.org}}
\begin{document}
\maketitle
\begin{abstract}
We investigate input-conditioned hypernetworks for multi-tasking in NLP, generating parameter-efficient adaptations for a decoder using a hypernetwork conditioned on the output of an encoder.
This approach produces a unique decoder adaptation for every input instance, allowing the network a larger degree of flexibility than prior work that only produces one decoder adaptation per task.
We apply our method to sequence classification tasks, extractive QA, and summarisation and find that it surpasses previous parameter efficient fine-tuning methods and often outperforms fully finetuning the underlying model. 
An analysis of the embeddings used by our hypernetwork shows that they are sensitive to output label and type, suggesting that our approach better maps from encoder representations to output labels. Our code is publicly available at \small{\texttt{\href{https://github.com/allenai/hyperdecoders}{https://github.com/allenai/hyperdecoders}}}.
\end{abstract}

\section{Introduction}

Recent work in NLP has examined the performance of large pretrained transformer-based models in multi-task settings, where a single model is evaluated on multiple tasks simultaneously, often with tasks converted to a shared sequence-to-sequence format \citep{2020t5, gpt3}. This greatly simplifies model training and deployment, requiring only one deployed model and format to handle multiple tasks. Additionally, training across multiple tasks can result in greatly improved performance for similar tasks \citep{stilts}, as well as tasks not seen during training \citep{sanh2022multitask, wei2022finetuned}.
However, not all tasks work well together, and jointly training on certain task pairs can reduce performance on both (`negative transfer') \citep{aribandi2022ext}.

In this paper, we propose a new method for multi-task NLP using a hypernetwork to generate an instance-specific decoder from the output of an encoder. We effectively explore if a model can learn to adapt \textit{itself} through learning how to generate adapter layers.
Our approach produces significant gains over prior approaches for efficient multi-task fine-tuning, often matching or exceeding full fine-tuning the underlying model. 

\begin{figure}
    \centering
    \begin{adjustbox}{width=.5\textwidth}
    \includegraphics{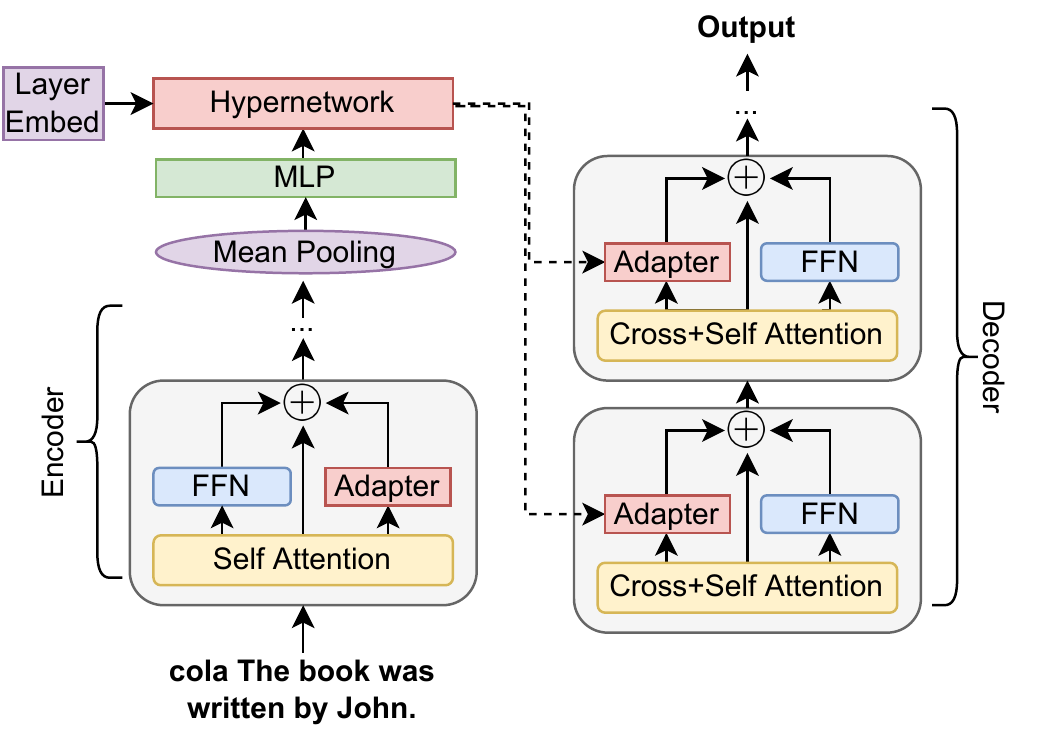}
    \end{adjustbox}
    \caption{An overview of our proposed approach, where a hypernetwork generates the adapters for the decoder in an encoder-decoder model.
    Given an input instance and task name, an encoder produces an embedding which is used to generate decoder adapter parameters using a hypernetwork.}
    \label{fig:main_fig}
\end{figure}

We build on parameter-efficient learning methods, where one trains a small set of parameters within a much larger model. These parameters may be newly introduced \citep{pmlr-v97-houlsby19a, li-liang-2021-prefix} or already exist within the model \citep{bitfit}, and are kept as few as possible. This means these methods often do not contain the capacity to learn multiple tasks at once, losing potential transfer benefits. One way to remain parameter-efficient while still handling a variety of tasks is to instead learn to generate these parameters, making use of an auxiliary network (`hypernetwork') to generate the weights used during inference \citep{tay2021hypergrid, pilault2021conditionally, ye-ren-2021-learning, karimi-mahabadi-etal-2021-parameter}. This allows the model to benefit from positive transfer between tasks through the shared hypernetwork while reducing negative transfer by allowing the generated parameters to be unique per task.

However, hypernetwork-based approaches generally condition their parameters on a learnt task embedding, meaning (a) the model is the same for every example within a task, and (b) adapting to new tasks requires further training to learn new task embeddings. We increase the flexibility of this approach by instead \textit{generating unique parameters for every input}, allowing the model to make use of similarities between samples across datasets and avoid potential interference between samples within the same dataset. This is achieved by using a shared encoder across all tasks and then generating adapter layers for the decoder only by feeding the encoded inputs into a hypernetwork. Furthermore, by conditioning on inputs rather than task embeddings, our approach allows simple transfer to out-of-domain data, as the shared hypernetwork and encoder learn to map from text to parameters. Figure \ref{fig:main_fig} illustrates our approach.

We apply our approach to a diverse set of tasks, including sequence classification, extractive question answering, and summarisation, and find that our approach outperforms existing parameter-efficient approaches and matches or outperforms full-finetuning. An analysis of our approach shows that sharing parameters in the encoder, but generating them in the decoder is more effective than other possible setups, suggesting that the encoder benefits from multi-task training while the decoder does not. Our results suggest that pretrained encoders can be easily adapted and trained to produce adaptations that enhance multi-task transfer learning for decoders, while producing useful adaptations for encoders is much more difficult.

To summarise, our core contributions are:
\begin{enumerate}
    \item We propose a new method for parameter-efficient multi-tasking, generating unique decoder layers for every input into a model.
    \item We show that our approach performs strongly against other parameter-efficient baselines and fully finetuning the underlying model across a diverse set of NLP tasks.
    \item We show the embeddings learnt by our hypernetwork are sensitive to both dataset and output label, suggesting the hypernetwork is effectively controlling the decoder.
\end{enumerate}

\section{Encoder-conditioned Decoders}

We make use of T5 as the underlying model in our experiments, which is a popular encoder-decoder model for sequence-to-sequence multi-tasking \citep{2020t5} and a common starting point for previous hypernetwork-based approaches \citep{karimi-mahabadi-etal-2021-parameter, tay2021hypergrid}. However, our overall approach can be applied to generic encoder-decoder transformer models.

\subsection{Adapter Layers}
We augment our underlying model with adapter layers \citep{pmlr-v97-houlsby19a}. These are small bottleneck networks with the following form:
\vspace{-.1cm}\begin{align}
    \text{Adapter}(\mathbf{x})= \mathbf{W}_{u} (f(\mathbf{W}_d \mathbf{x} + \mathbf{b}_d)) + \mathbf{b}_u
\end{align}

\vspace{-.1cm}Where $f$ is the ReLU activation function. We insert these layers in parallel with the feedforward module of each layer of a larger pretrained model, following \citet{he2022towards}:
\vspace{-.1cm}\begin{align}
   \mathbf{y} & = \text{FF}(\text{LayerNorm}(\mathbf{x})) + \text{Adapter}(\mathbf{x})
\end{align}

\vspace{-.1cm}Where $\mathbf{y}$ is the output for the layer, FF is the feedforward module, and $\mathbf{x}$ is the output from the attention module(s). We find using $\text{LayerNorm}(\mathbf{x})$ as input to the adapter less effective than directly using $\mathbf{x}$ (see section \ref{sec:ablations}). During training, the underlying network is frozen and only $\mathbf{W}_{u}$, $\mathbf{W}_{d}$, $\mathbf{b}_{u}$, and $\mathbf{b}_{d}$ are updated.

\subsection{Adapter-generating Hypernetworks}
\label{sec:hypernetwork}

A hypernetwork is a network that produces the parameters used for another network \citep{hypernetworks, fwp1991}. We use a simple two-layer network to produce adapter parameters $\mathbf{W}_d$, $\mathbf{W}_u$, $\mathbf{b}_d$, $\mathbf{b}_u$. Given some input $x$ to the hypernetwork, we generate these parameters as follows:
\vspace{-.1cm}\begin{align}
    \mathbf{h} & = \text{ReLU}(\mathbf{W}_0\mathbf{x} + \mathbf{b}_0) \\
    \mathbf{W}_u & = \mathbf{W}_1\mathbf{h} + \mathbf{b}_1 \\
    \mathbf{W}_d & = \mathbf{W}_2\mathbf{h} + \mathbf{b}_2 \\
    \mathbf{b}_u & = \mathbf{W}_3\mathbf{h} + \mathbf{b}_3 \\
    \mathbf{b}_d & = \mathbf{W}_4\mathbf{h} + \mathbf{b}_4
\end{align}

\vspace{-.1cm}We re-use this hypernetwork to generate the adapters for every layer by (partially) conditioning the input on layer embeddings, greatly improving the parameter efficiency of this approach.
The hypernetwork parameters are initialised using the method proposed in \citet{Chang2020Principled}.

\subsection{Encoder-conditioned Decoders}

The core idea of our approach is to condition a hypernetwork on the output of the encoder to generate the adapters used for the decoder in an encoder-decoder model. We place regular (non-generated) adapter layers in the encoder to allow it to adapt to tasks in a parameter-efficient way. We then feed the encoder output to a hypernetwork to generate custom decoder adapters for every input to the encoder. This allows our approach to flexibly differentiate between samples within a task, rather than remaining static per task. This potentially allows more flexible transfer learning both between samples within datasets and with samples across datasets. We explore other potential configurations in section \ref{sec:ablations} and find ours works best overall. We name this approach `hyperdecoder' in the following experiments.

Concretely, given some input, we first pass it through the T5 encoder to construct a hidden representation $\mathbf{h}$. The T5 encoder is equipped with adapter layers, which are trained during fine-tuning while the rest of the encoder is kept frozen. We mean-pool this representation and pass it through a two-layer network with a ReLU activation to construct a vector embedding of the input:
\vspace{-.1cm}\begin{equation}
    \mathbf{e} = \text{MLP}(\text{mean}(\mathbf{h}))
    \label{embed}
\end{equation}

\vspace{-.1cm}We then generate the parameters for an adapter in each layer $i$ of the decoder by concatenating a learnt layer embedding $\mathbf{l}_i$ to the embedding and passing it through the hypernetwork described in section \ref{sec:hypernetwork}.
\vspace{-.1cm}\begin{align}
    \text{Adapter}_i & = \text{Hypernetwork}([\mathbf{e}; \mathbf{l}_i])
\end{align}

\vspace{-.1cm}All parameters in the decoder are frozen and the hypernetwork is trained along with the encoder adapters during fine-tuning. This approach is summarised in Figure \ref{fig:main_fig}.

\subsection{Multi-tasking}

We focus on a multi-task setup, where a model is trained on multiple tasks simultaneously. The underlying model parameters $\theta$ are kept unchanged during training and only the adapter and hypernetwork parameters $\theta'$ are updated. All parameters are shared between all tasks, and the model is trained in a seq2seq setting with cross-entropy loss, following \citet{2020t5}.

\section{Experiments}
This section details our experimental setup and baselines.  Results are given in Section \ref{sec:results}.

\begin{table*}
\centering
\begin{adjustbox}{width=1\textwidth,center=\textwidth}
\begin{tabular}{l|c|llllllll|l}
\hline
Model & \begin{tabular}{@{}c@{}}\% Trainable \\ Param.\end{tabular} & CoLA & SST-2 & STS-B & MRPC & QQP & MNLI & QNLI & RTE & Avg\\
\hline
Full Finetuning & 100\% & \textbf{63.6} & 94.8 & \textbf{92.0} / 91.6 & 88.7 / 91.8 & \textbf{92.2 / 89.5} & 88.6 & 93.3 & 77.5 & 86.3 \\
Hyperformer & 8.8\% & 19.2 & 87.3 & 86.2 / 85.8 & 73.4 / 81.3 & 87.0 / 82.8 & 77.7 & 84.2 & 55.1 & 71.5 \\
Hyperformer++ & 4.6\% & 35.5 & 88.3 & 87.9 / 87.6 & 78.8 / 85.2 & 86.4 / 82.3 & 88.3 & 84.8 & 64.5 & 76.9 \\
Task Hypernet & 2.7\% & 0.0 & 82.1 & 16.4 / 16.4 & 70.4 / 81.4 & 89.8 / 86.5 & 56.6 & 64.8 & 50.7 & 54.3 \\
Modular Hypernet & 4.8\% & 51.2 & 96.2 & 91.9 / \textbf{92.0} & \textbf{90.1 / 93.0} & 89.4 / 86.1 & 89.5 & 93.5 & 74.6 & 84.5 \\
Adapter & 2.9\% & 58.5$_{3.1}$ & 95.7$_{0.3}$ & 90.1$_{1.8}$ / 90.3$_{1.9}$ & 89.4$_{1.5}$ / 92.2$_{1.0}$ & 91.4$_{1.9}$ / 88.6$_{0.3}$ & 89.8$_{0.1}$ & 94.1$_{0.2}$ & 80.7$_{1.8}$ & 86.2$_{0.6}$ \\
\textbf{Hyperdecoder (ours)} & 2.9\% & 58.7$_{2.3}$ & \textbf{95.9}$_{0.4}$* & 91.8$_{0.7}$ / \textbf{92.0}$_{0.4}$* & 89.2$_{1.5}$ / 92.0$_{0.9}$ & 91.1$_{0.2}$ / 88.3$_{0.4}$ & \textbf{90.0}$_{0.2}$* & \textbf{94.2}$_{0.4}$ & \textbf{80.8}$_{2.2}$ & \textbf{86.5}${_{0.5}}\dagger$ \\ \hline
\end{tabular}
\end{adjustbox}
\caption{Performance of models using \textbf{T5$_{\text{large}}$ v1.1 + LM} as the base model on GLUE test splits described in section \ref{sec:data_glue}.  Trainable parameters is \% of parameters trained in the model compared to fully finetuning T5. We report mean and standard deviation over 25 runs for Adapter and Hyperdecoder. * indicates value is statistically significant ($p < 0.05$). $\dagger$ Hyperdecoder is statistically significantly better when CoLA is removed from average.}
\label{tab:glue_11}
\end{table*}

\subsection{Datasets}
We evaluate our approach in three settings: GLUE \citep{wang-etal-2018-glue}, the 2019 MRQA shared task \citep{fisch-etal-2019-mrqa}, and a set of summarisation and NLI datasets. For each setting, we sample from sub-tasks proportionally to their size, as initial experiments showed more complex sampling techniques provided little to no benefit. All tasks are in English.

\subsubsection{GLUE}
\label{sec:data_glue}
GLUE \citep{wang-etal-2018-glue} is a set of sequence classification tasks including paraphrase detection \cite[QQP and MRPC;][]{dolan2005automatically},
semantic similarity \cite[STS-B;][]{agirre2007semantic},
natural language inference \cite[MNLI;][]{williams2018broad},
\cite[QNLI;][]{rajpurkar2016squad}, \cite[RTE;][]{dagan2006pascal, bar2006second, giampiccolo2007third, bentivogli2009fifth}, linguistic acceptibility \cite[CoLA;][]{warstadt2018neural}, and sentiment classification \cite[SST-2;][]{socher2013recursive}. Following  \citet{karimi-mahabadi-etal-2021-parameter}, for datasets with small training sets (RTE, MRPC, STS-B, CoLA), we split the validation set in half into test and validation sets, for other larger datasets we split out 1000 examples to use as the validation set and use the original validation set as a test set, and for MNLI we use the mismatched validation set as the test set. Following prior work
we do not evaluate on WNLI. We preprocess the GLUE inputs to follow the format used by \citet{2020t5} (including task prefix).

\subsubsection{MRQA}

The MRQA 2019 shared task dataset \citep{fisch-etal-2019-mrqa} is a collection of 12 QA datasets, all modified to the same format of extractive QA. 6 datasets are used for training and evaluation: HotpotQA \cite{yang-etal-2018-hotpotqa}, Natural Questions \cite{kwiatkowski-etal-2019-natural}, NewsQA \cite{trischler-etal-2017-newsqa}, SQuAD \cite{rajpurkar2016squad}, SearchQA \cite{searchqa}, and TriviaQA \cite{joshi-etal-2017-triviaqa}.
Another 6 are used for out-of-domain evaluation: BioASQ \cite{bioasq}, DROP \cite{dua-etal-2019-drop}, DuoRC \cite{saha-etal-2018-duorc}, RACE \cite{lai-etal-2017-race}, RelationExtraction \cite{levy-etal-2017-zero}, and TextbookQA \cite{Kembhavi_2017_CVPR}. This effectively tests a model's ability to generalise to out-of-domain data. We evaluate on the validation split of all 12 datasets after all training steps are complete. We preprocess all MRQA data to follow the SQuAD template used by \citet{2020t5}\footnote{`\texttt{question: <question> context: <context>}'.}.  Notably, this prompt does not provide any indication as to which dataset a given input belongs. We split contexts into chunks of length 512 tokens with an overlap of 128 tokens. We pair chunks with answers with the answer found in the chunk, and chunks without answers with empty strings. At evaluation time, we produce an answer for all chunks and take the most likely non-empty string as the final answer.

\begin{table*}
\centering
\begin{adjustbox}{width=1\textwidth,center=\textwidth}
\begin{tabular}{l|c|llllllll|l}
\hline
Model & \begin{tabular}{@{}c@{}}\% Trainable \\ Param.\end{tabular} & CoLA & SST-2 & STS-B & MRPC & QQP & MNLI & QNLI & RTE & Avg\\\hline
Hyperformer* & $\sim$4\% & \textbf{58.9} & 95.7 & 91.6 / 91.5 &  92.7 / 90.0 & 87.7 / 90.7 & 89.8 & 94.5 & 87.0 & 87.3 \\
HyperPrompt* & $\sim$4\% & 57.5 & \textbf{96.7} & \textbf{91.9 / 92.0} &  93.6 / 91.2 & 87.0 / 90.1 & 90.3 & \textbf{95.0} & 87.7 & 87.5 \\
\textbf{Hyperdecoder (ours)} & 4.2\% & 58.2 & 96.4 & 91.5 / 91.6 & \textbf{93.8 / 91.4} & \textbf{89.3 / 91.9} & \textbf{90.7} & 94.8 & \textbf{88.4} & \textbf{87.9} \\
\hline
\end{tabular}
\end{adjustbox}
\caption{Performance of models using \textbf{T5$_{\text{large}}$ v1.1 + LM} as the base model on GLUE dev set splits, picking the best task performance across all checkpoints.  Trainable parameters is \% of the trainable parameters used compared to fully finetuning T5. * Results from \citet{He2022HyperPromptPT}.}\label{tab:glue_hyperprompt}
\end{table*}
\begin{table*}
\centering
\begin{adjustbox}{width=1\textwidth,center=\textwidth}
\begin{tabular}{l|c|llllllll|l}
\hline
Model & \begin{tabular}{@{}c@{}}\% Trainable \\ Param.\end{tabular} & CoLA & SST-2 & STS-B & MRPC & QQP & MNLI & QNLI & RTE & Avg\\
\hline 
Full Finetuning* & 100\% & 54.9 & 92.5 &  88.8 / 88.5 & 90.2 / 93.0 & \textbf{91.1 / 88.1} &  85.7 & 92.0 & 75.4 & 83.8 \\
Hyperformer* & 54\% & 61.3 & 93.8 & 89.6 / 89.0 & \textbf{90.6 / 93.3} &  87.2 / 90.1 & \textbf{86.3} & 92.8 & \textbf{78.3} & \textbf{85.3} \\
Hyperformer++* & 2\% & \textbf{63.7} & \textbf{94.0} & 90.0 / 89.7 & 89.7 / 92.6 & 87.2 / 90.3 & 85.7 & 93.0 & 75.4 & 85.2 \\ 
\textbf{Hyperdecoder (ours)} & 7.4\% & 54.8 & 93.8 & \textbf{90.3 / 90.2} & 86.2 / 90.5 & 90.5 / 87.3 & 85.8 & \textbf{93.4} & 71.0 & 83.3 \\ \hline
\end{tabular}
\end{adjustbox}
\caption{Performance of models using \textbf{T5$_{\text{base}}$ vanilla} as the base model on GLUE test splits described in section \ref{sec:data_glue}. * Results reported by \citet{karimi-mahabadi-etal-2021-parameter}.}\label{tab:glue_10} 
\end{table*}

\subsubsection{Summarisation and NLI}
\label{sec:data_xsum_nli}
To investigate transfer learning with difficult tasks, we evaluate on summarisation and NLI tasks that are known to cause negative interference  \citep{aribandi2022ext} and are difficult for current parameter-efficient techniques \citep{he2022towards}. For summarisation, we use Xsum \citep{narayan-etal-2018-dont}, CNN/Daily Mail \citep{NIPS2015_afdec700}, and the English WikiLingua split from \citet{gehrmann-etal-2021-gem}. For NLI, we use MNLI \citep{williams2018broad}, abductive NLI \citep{bhagavatula2020abductive}, and adversarial NLI \citep{nie-etal-2020-adversarial}. We jointly train and evaluate on all tasks. We preprocess all datasets following the templates used in \citet{2020t5}\footnote{This means that the NLI tasks have their dataset name prepended onto the prompt, while summarisation tasks do not.}. We evaluate on the provided test splits for all datasets except abductive NLI and MNLI. For abductive NLI, we split 1000 samples from the validation set and use the existing validation set as the test set. We treat MNLI as detailed in section \ref{sec:data_glue}.

\subsection{Experimental Details}
\label{sec:exp_details}
We build on the Hyperformer codebase \citep{karimi-mahabadi-etal-2021-parameter}, making use of the transformers implementation of T5 \citep{wolf-etal-2020-transformers}. We finetune all models using AdamW \citep{loshchilov2018decoupled}, with a learning rate of 3e-4 with linear decay and 500 warmup steps. For GLUE tasks, we train for 65k steps with an effective batch size of 128, evaluate every 1000 steps on the development set, and test on the overall best performing checkpoint. For MRQA, we train for 4 epochs and evaluate on the final model (initial experiments showed that taking checkpoints always resulted in evaluating on the final model regardless). For summarisation and NLI tasks, we train for 100k steps with a batch size of 64, evaluate every 5000 steps on the development set, and test using the single overall best-performing checkpoint. All non-hypernetwork and non-adapter parameters are frozen throughout training. All experiments start from the T5 v1.1+LM checkpoints \citep{lester-etal-2021-power} unless otherwise stated. Further details can be found in Appendix \ref{sec:training_details}.

\subsection{Baselines}

We primarily compare against three strong baselines: fully-finetuning the underlying model, training only adapter layers placed in parallel with all feedforward modules (`adapter') and using task-conditioned hypernetworks to generate adapter layers in the encoder and decoder (`task hypernet'), similar to the Hyperformer \citep{karimi-mahabadi-etal-2021-parameter}. Apart from full finetuning, we keep the number of trainable parameters roughly equal across methods. More details are provided in Appendix \ref{sec:expp_details}.

We additionally compare against relevant prior work. For GLUE, we compare against the Hyperformer and Hyperformer++ models proposed in \citet{karimi-mahabadi-etal-2021-parameter} with an increased number of trainable parameters\footnote{We increased the number of trainable parameters for Hyperformer as a best-faith effort to provide a strong baseline as the default settings tended to result in worse performance in initial experiments.}, and the modular task hypernetwork \citep{ponti2022combining}. For MRQA, we compare against CA-MTL \citep{pilault2021conditionally}, which modifies a BERT model with several task-conditional modules and uses a novel sampling technique.  We also compare to UnifiedQA \citep{khashabi-etal-2020-unifiedqa} results reported by \citet{friedman-etal-2021-single} to show out-of-domain split performance from alternate T5-based QA model.

\section{Results}
\label{sec:results}
This section details our main experimental results.  Ablations and analysis follow in Section \ref{sec:analysis}.

\subsection{GLUE}
\label{sec:exp_glue}

We report our results on the GLUE benchmark in Tables \ref{tab:glue_11} and \ref{tab:glue_10}. Our approach improves greatly over the Hyperformer and full finetuning when using T5 v1.1 + LM as the underlying model. Notable improvements are made in SST-2 and RTE tasks, the latter of which is known to benefit from transfer learning \citep{stilts}. This suggests our approach enhances positive transfer benefits over adapter-only and full finetuning approaches. The worst performing approach is the `task hypernet' method, which follows the Hyperformer approach but with our adapter placement. This suggests that the adapter placement difference between our method and the Hyperformer is not the reason for our improved performance, but rather the use of an encoder-conditioned decoder. Additionally, our approach remains parameter efficient, training 0.03$\times$ fewer parameters than full finetuning. Our approach also outperforms HyperPrompt \citep{He2022HyperPromptPT} when using a matching evaluation setup\footnote{\citet{He2022HyperPromptPT} do not release their code, so we are limited to comparing against the numbers they report.}, as seen in \ref{tab:glue_hyperprompt}.

However, we note that in table \ref{tab:glue_10} our approach underperforms when using the original T5 model as the underlying model. As T5 was originally pretrained with a mix of self-supervised span-infilling and supervised tasks (including GLUE), this suggests the Hyperformer is able to effectively adapt the model to tasks seen or similar to those seen during pretraining. However, when we remove these tasks from the pretraining mixture (as was done for T5 v1.1+LM), the Hyperformer struggles to adapt the model, as seen in Table \ref{tab:glue_11}. Overall, this suggests being exposed to the underlying task in pretraining can make a large difference in the evaluation of parameter-efficient methods. As we wish to evaluate how well our approach can adapt models to completely unseen tasks, we restrict our underlying model to T5 v1.1+LM.

\subsection{MRQA}
\label{sec:mrqa_Res}

\begin{table*}[]
\begin{adjustbox}{width=1\textwidth,center=\textwidth}
\begin{tabular}{lcccccc|c}
\hline
Model & SQuAD & HotpotQA & TriviaQA & NewsQA & SearchQA & Natural Qs & Avg \\ \hline
BERT-base & 86.7 & 76.6 & 71.6 & 66.8 & 76.7 & 77.4 & 62.6 \\
BERT-large & 88.4 & 79.0 & 74.7 & 66.3 & 79.0 & \textbf{79.8} & 77.9 \\
CA-MTL (BERT-large)* & - & - & - & - & - & - & 79.5 \\\hline
Full Finetuning & 91.0 & \textbf{80.6} & \textbf{76.2} & \textbf{69.0} & \textbf{83.4} & 79.6 & \textbf{80.0} \\
Adapter & 90.8 & 79.5 & 75.0 & 68.6 & 82.9 & 78.6 & 79.2 \\
Task Hypernet & 88.9 & 77.4 & 70.6 & 66.0 & 81.0 & 75.0 & 76.5 \\
\textbf{Hyperdecoder (ours)} & \textbf{91.3} & 79.9 & 75.0 & 68.6 & 82.9 & 79.1 & 79.5 \\ \hline
\end{tabular}
\end{adjustbox}
\caption{F1 score of models on in-domain MRQA validation split. * results from \citet{pilault2021conditionally}, who do not report performance on individual datasets and train on additional data. All non-BERT models use T5$_{\text{base}}$ v1.1 + LM.}\label{tab:in_mrqa}
\end{table*}

\begin{table*}[]
\begin{adjustbox}{width=1\textwidth,center=\textwidth}
\begin{tabular}{lcccccc|c}
\hline
Model & BioASQ & DROP & DuoRC & RACE & Relation Ext. & TextbookQA & Avg \\ \hline
BERT-base & 62.7 & 34.5 & 54.6 & 41.4 & 83.8 & 53.9 & 55.2 \\
BERT-large & 66.8 & 43.8 & 58.0 & 42.5 & 85.2 & 55.7 & 58.7\\
UnifiedQA* & 59.7 & 45.7 & 30.4 & 51.4$\dagger$ & 82.0 & 35.9 & 50.9\\\hline
Full Finetuning & 63.9 & 47.9 & 54.5 & 47.3 & 84.0 & 55.1 & 58.8 \\
Single Adapter & \textbf{66.0} & \textbf{48.2} & 55.6 & \textbf{47.7} & 84.0 & 57.5 & 59.8 \\
Task Hypernet** & 63.9 & 36.6 & 53.0 & 44.8 & 82.0 & 53.0 & 55.6 \\
\textbf{Hyperdecoder (ours)} & 65.8 & 47.0 & \textbf{58.1} & 46.3 & \textbf{84.3} & \textbf{59.5} & \textbf{60.2} \\ \hline
\end{tabular}
\end{adjustbox}
\caption{F1 score of models on out-of-domain MRQA validation split. All non-BERT models use T5$_{\text{base}}$ v1.1 + LM. * results from \citet{friedman-etal-2021-single}. **Task Hypernet uses an average of the learnt in-domain task embeddings to condition its adapters. $\dagger$ RACE was part of UnifiedQA's training data.}\label{tab:out_mrqa}
\end{table*}

We report our results on the MRQA dataset in Tables \ref{tab:in_mrqa} and \ref{tab:out_mrqa}. We note that during experimentation, we found it beneficial to increase the encoder adapter size and reduce the decoder adapter size, keeping the overall parameter budget roughly the same (full results detailed in Appendix \ref{sec:mrqa_full_results}). Overall, our approach outperforms other parameter-efficient approaches and full finetuning (69.4 vs 69.9 overall average F1 between our method and full-finetuning). Gains are especially large in the out-of-domain test sets, likely due to the fact that freezing the underlying model preserves knowledge useful to these new domains. We also note our method performs especially well on long-context out-of-domain datasets (DuoRC, TextbookQA), suggesting it is especially effective at identifying when the question cannot be answered given a subsection of context. This shows our approach is still able to work well even when the underlying datasets are not revealed to the model in the prompt and that it can generalise well to out-of-domain data. Additionally, our approach matches CA-MTL, which uses an underlying model with approximately 90 million extra parameters (BERT-large vs T5$_\text{base}$) and makes use of additional training data. Overall, this suggests our approach is able to generalise well to out-of-domain data, even when underlying datasets are not distinguished.

\subsection{Summarisation \& NLI}
\label{sec:exp_sum}

\begin{table}[]
\begin{adjustbox}{width=.5\textwidth}
\centering
\begin{tabular}{lccc}
\hline
Model & Sum. Avg. & NLI Avg. & Overall Avg. \\ \hline
Full-finetuning & \textbf{18.4} & 64.8 & 41.6 \\
Adapter & 17.0 & \textbf{66.3} & \textbf{41.7} \\
Task Hypernet & 14.0 & 62.0 & 38.0 \\
\textbf{Hyperdecoder (ours)} & 16.8 & 65.6 & 41.2 \\ \hline
\end{tabular}
\end{adjustbox}
\caption{Average performance across Summarisation and NLI datasets using T5$_{\text{base}}$ v1.1 + LM. Summarisation performance is average of R2 scores, while NLI is average of accuracy scores. Overall average is the arithmetic mean of the two.}\label{tab:xsum_nli}
\end{table}

Finally, to explore a setting where strong negative interference is present, we experiment on a combination of summarisation and NLI tasks following \citet{aribandi2022ext}. As shown in Table \ref{tab:xsum_nli}, we find that while no approach is able to match full-finetuning in summarisation, both regular adapters and our approach are able to perform well for NLI. This suggests that while our approach is able to avoid some of the negative interference that results in lower NLI scores for the fully-finetuned model, it still struggles to overcome it. We note that similar to MRQA, we found here that placing a larger parameter budget into the encoder outperformed evenly splitting the budget between encoder and decoder (see Appendix \ref{sec:xsum_nli_full_results} for details).

\section{Analysis}
\label{sec:analysis}
This section presents an analysis of the Hyperdecoder's embeddings and ablations, establishing the efficacy of our architecture design choices.

\subsection{Hypernetwork Embeddings}

\begin{figure}
    \centering
    \begin{adjustbox}{width=.45\textwidth}
    \includegraphics{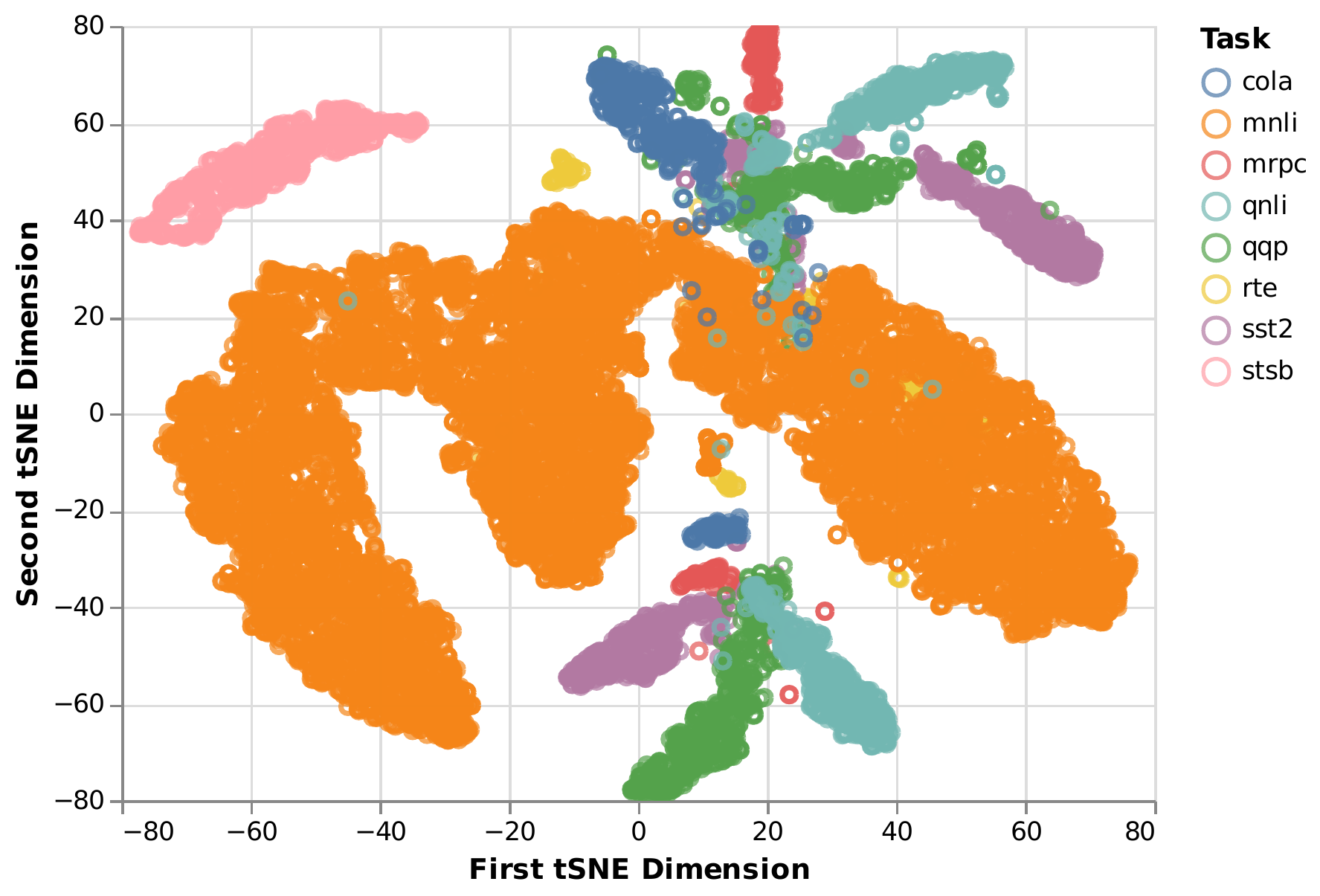}
    \end{adjustbox}
    \caption{t-SNE visualisation of GLUE validation set hypernetwork embeddings produced by our approach.}
    \label{fig:glue_embed}
\end{figure}

\begin{figure}[t]
    \centering
    \begin{adjustbox}{width=.45\textwidth}
    \includegraphics{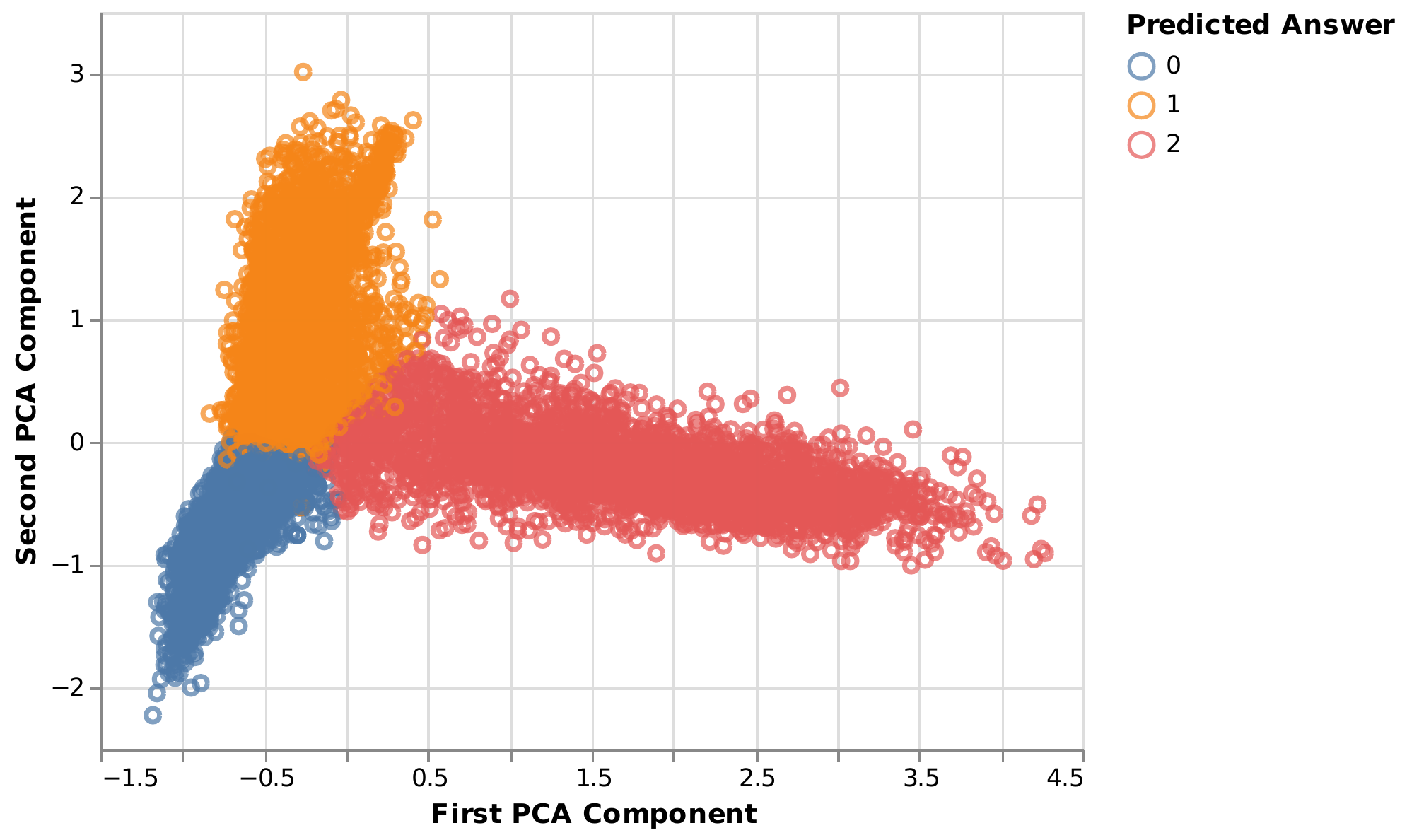}
    \end{adjustbox}
    \caption{PCA visualisation of GLUE validation set hypernetwork embeddings, coloured by predicted label. STS-B examples removed for simplicity, as it is cast to a 21-class classification task for T5.}
    \label{fig:glue_embed_labels}
\end{figure}

\begin{figure}
    \centering
    \begin{adjustbox}{width=.45\textwidth}
    \includegraphics{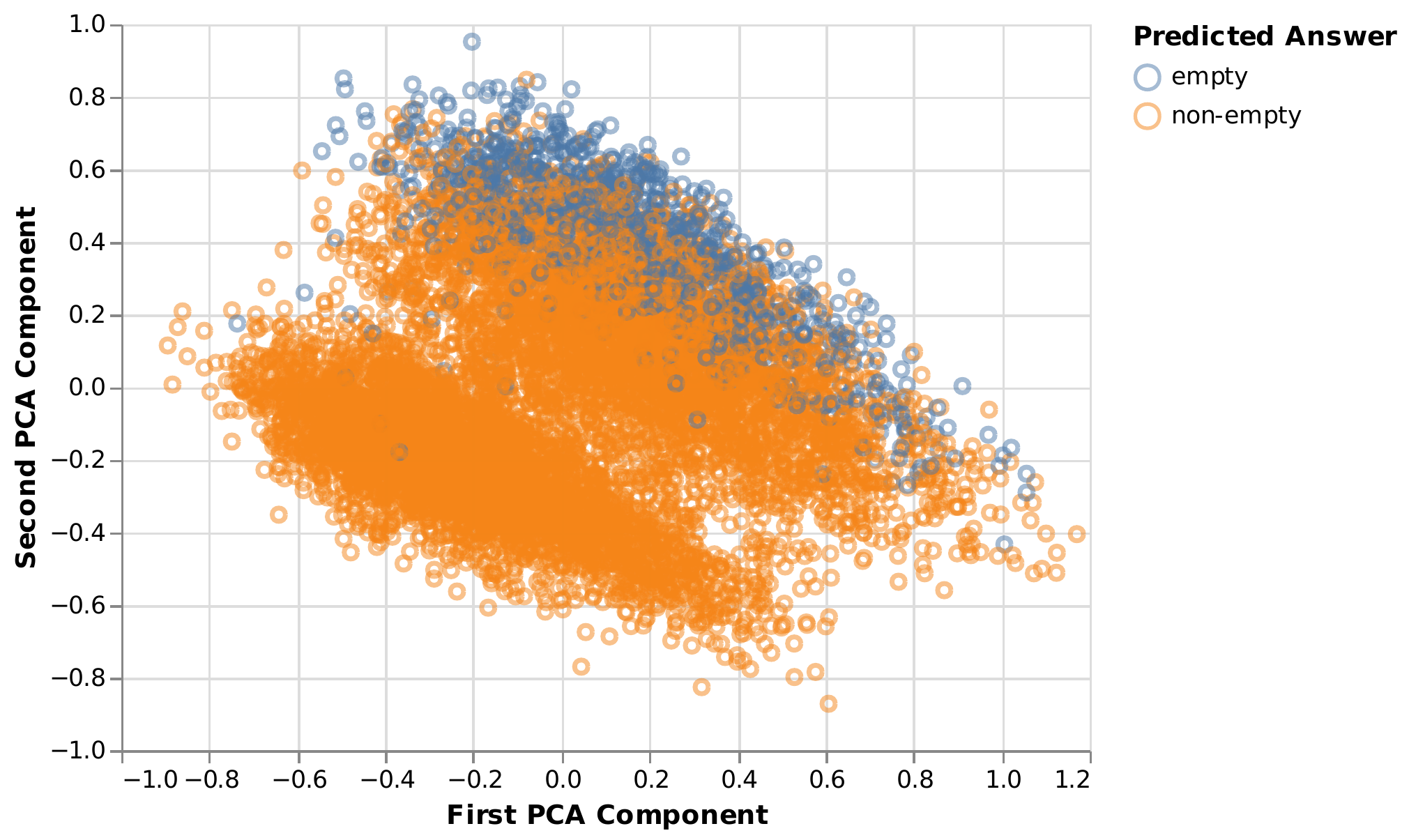}
    \end{adjustbox}
    \caption{PCA visualisation of hypernetwork embeddings for NewsQA instances. Blue indicates predicted answer is the empty string, orange non-empty string.}
    \label{fig:newsqa_embed}
\end{figure}

We visualise the embeddings learnt by the encoder before being passed to the hypernetwork (`$\mathbf{e}$' in Equation \ref{embed}) by utilising dimensionality reduction with t-SNE \citep{tsne} and PCA \citep{pca}. Although Figure \ref{fig:glue_embed} suggests the hypernetwork does customise to different datasets to a degree, Figure \ref{fig:glue_embed_labels} shows the most salient difference between samples is the predicted label. Note that following \citet{karimi-mahabadi-etal-2021-parameter} we train our models to output numeric rather than text labels, meaning that while the labels may have semantic differences between datasets, the actual output from the decoder is identical between datasets. This suggests the hypernetwork has learnt to map from embedding space to the text labels, and the encoder is doing much of the classification work. We further investigate this by training simple linear classifiers for all datasets apart from STS-B on top of the T5 encoder (with learnt adapters) in Table \ref{tab:enc_only_glue}. We note that we can recover much of the performance of our model in this case, suggesting that the decoder is largely working to map from represented space to text label space. Furthermore, the efficacy of our model at long-context out-of-domain datasets for MRQA further suggests that the hypernetwork can effectively control the decoder to output specific labels when needed, \textit{but can flexibly swap to generate arbitrary text output when needed}, unlike a simple linear classifier. This is especially important for multi-tasking and long-context documents where the model must swap between generating short set labels and arbitrary longer text. A visualisation of the hypernetwork embeddings generated for NewsQA in Figure \ref{fig:newsqa_embed} further shows that empty string answers are generally clustered together.

\begin{table}[t]
\begin{adjustbox}{width=.5\textwidth}
\centering
\begin{tabular}{lccc}
\hline
Model                  & Encoder-decoder & Encoder-only \\ \hline
\textbf{Hyperdecoder (ours)} & \textbf{75.23} & 74.94 \\
Adapter & 74.69 & 72.65 \\ \hline
\end{tabular}
\end{adjustbox}
\caption{GLUE validation set performance for encoder-only and encoder-decoder models excluding STS-B. We detail the encoder-only setup in Appendix \ref{sec:encoder_only_expl}.}\label{tab:enc_only_glue}
\end{table}

\subsection{Ablations}
\label{sec:ablations}

\paragraph{Placement of adapter and parameter generators}

\begin{table*}[]
\centering
\begin{adjustbox}{width=1\textwidth,center=\textwidth}
\begin{tabular}{l|cccc}
\hline
Encoder$\downarrow$\hfill Decoder$\rightarrow$ & Full-Finetuning & Adapters & Task-Conditioned Adapters & Encoder-Conditioned Adapters \\ \hline
Full-Finetuning & 86.3 & - & - & - \\
Adapters & - & 86.4 & 86.3 & \textbf{86.9} \\
Task-Conditioned Adapters & - & 51.5 & 54.3 & 57.9 \\
Encoder-Conditioned Adapters & - & 83.4 & 82.3 & 82.8 \\ \hline
\end{tabular}
\end{adjustbox}
\caption{Average GLUE benchmark performance across varied encoder/decoder adapter configurations using T5$_\text{large}$ v1.1 + LM as the base model. Number of trainable parameters is kept similar across approaches apart from full-finetuning. More details given in Appendix \ref{sec:glue_ablation_full_results}.}\label{tab:glue_ablation}
\end{table*}

We investigate alternate adapter generation possibilities by varying regular, task, and encoder-conditioned adapters independently in the encoder and decoder while keeping the total number of trainable parameters roughly constant. We use the same setup as our previous GLUE experiments. As seen in Table \ref{tab:glue_ablation}, input-conditioned or task-conditioned adapters do not perform as well as regular adapters in the encoder. The task hypernetwork struggles to learn useful adapters at all, while the encoder-conditioned adapters perform better but still do not match directly learnt adapters, likely due to the regular adapters being more effectively able to share knowledge or simply being easier to optimise. However, it seems much easier to learn to generate adapters for the decoder, with both task and encoder-conditioned adapters performing well. Our approach, using regular adapters in the encoder and encoder-conditioned adapters in the decoder, performs best overall.

\paragraph{Other Elements}

\begin{table}[]
\centering
\begin{tabular}{lccc}
\hline
Model & GLUE Avg \\ \hline
Hyperdecoder & \textbf{86.9} \\
- MLP & 86.7 \\
+ post-layernorm input & 83.2 \\ \hline
\end{tabular}
\caption{GLUE performance over different ablations.}\label{tab:other_ablations}
\end{table}

We also investigate removing the MLP used in Equation \ref{embed} and using layer normalisation outputs as input to the adapters (`post-layernorm input'). The results in Table \ref{tab:other_ablations} suggest the MLP provides some utility and using inputs pre-layer normalisation works better.

\section{Related Work}

\subsection{Multi-task Models}

Neural networks have long been known to uncover and make use of task relatedness when training across multiple tasks \citep{multitaskold}. Applications of this approach in NLP initially have usually involved generating shared representations and passing these to task-specific layers \citep{unifiednlpmodel, liu2019mt-dnn}. Newer approaches have opted to use a single set of parameters for all tasks, achieved by casting them to unified formats \citep{2020t5}. This allows massive multi-tasking approaches where extremely large models are trained across a wide variety of tasks \citep{aghajanyan-etal-2021-muppet, aribandi2022ext}, often with benefits to few or zero-shot performance \citep{sanh2022multitask, wei2022finetuned}. This requires finetuning large models for long amounts of time over a large number of tasks, which may be out of reach with a limited compute budget.

\subsection{Parameter-efficient Tuning}

Numerous parameter-efficient approaches to finetuning large models have been proposed, including adapters \citep{pmlr-v97-houlsby19a}, prefix-tuning \citep{li-liang-2021-prefix},  prompt-tuning \citep{lester-etal-2021-power}, and p-tuning \citep{ptuning2,liu2021gpt}, all of which involve learning a small set of parameters in carefully chosen locations to achieve performance close to fully-finetuning the model. Recent studies have shown the effectiveness of these methods can be increased with differing placements \citep{pfeiffer-etal-2021-adapterfusion, he2022towards} and that parameters learnt for one task or language can be combined to allow better performance across a wide variety of tasks or languages \citep{pfeiffer-etal-2021-adapterfusion, pfeiffer-etal-2020-mad}.

\subsection{Hypernetwork-based Adaption Methods}

Generating network weights with another network was originally proposed by \citet{fwp1991} in a general setting. More recently, \citet{hypernetworks} showed modulating weights in CNNs and LSTMs could improve performance on various tasks, including language modelling. With the rise of large pretrained transformers in NLP, much recent work has explored generating adaptations for these models. \citet{tay2021hypergrid} and \citet{karimi-mahabadi-etal-2021-parameter} investigate generating adapter (or adapter-like) layers using task embeddings and hypernetworks for multitasking, evaluating primarily on sequence classification tasks using T5 v1.0. \citet{ye-ren-2021-learning} investigate generating adapters from task descriptions for zero-shot tasks and find this provides improvements over just fully-finetuning the underlying model. Concurrently, \citet{volk2022example} use a T5 model to convert input instances into domain `signatures', which they condition a hypernetwork on to generate classifier weights. In contrast, we directly generate weights for adapters from the encoded input representation. Recent work has further applied hypernetwork-based adaptation methods to vision and language tasks \citep{Sung2021VLAdapterPT}, prompt-tuning \citep{He2022HyperPromptPT}, and few-shot-based multitasking \citep{ponti2022combining}.

Outside of sequence-to-sequence-based approaches, \citet{pilault2021conditionally} find modifying multiple parts of BERT to be conditional on a task embedding to be effective for sequence classification tasks. \citet{ustun-etal-2020-udapter} and \citet{ansell-etal-2021-mad-g} also explore generating multilingual adapters for mBERT using linguistic typological features, the former using parameter generators in every layer.

\section{Conclusion}

We propose a novel method for generating adapters conditioned on a model's input and show that this improves performance in multi-task settings across a variety of tasks. We explore the effectiveness of our approach for sequence classification, QA, and summarisation tasks, and find that it often outperforms strong parameter-efficient baselines. Future work could examine applying our approach to other architectures (e.g. decoder-only models) or explore the tradeoffs between shared and generated parameters across different layers. An analysis of our approach suggests the primary benefits come from improved control of the encoder over the decoder, enhancing the effects of positive transfer from the shared encoder. This allows our approach to efficiently adapt a pretrained language model to multiple tasks unseen during pretraining while still benefiting strongly from positive transfer.

\section*{Limitations}

Our work explores a novel idea within parameter-efficient finetuning by conditioning a model on itself, enabling greater flexibility for multi-tasking while adding relatively few parameters. However, this flexibility has limits: some tasks are more difficult to adapt with our method than others, as seen in the summarisation results in section \ref{sec:exp_sum}. Examining other parameter efficient training methods such as LoRA \citep{hu2022lora} or prefix-tuning \citep{li-liang-2021-prefix} on top of adapters may yield further improvements \citep{he2022towards}, but may also uncover a dependence on the particular adapter setup used by our approach. Additionally, parts of our design involve compressing information (in particular, the mean-pooling used to condition the hypernetwork), and further experiments on tasks with long inputs or outputs (such as summarisation) may reveal potential limitations of this approach and suggest further improvements. Additionally, our work only examines English-based tasks, so the ability of the model to handle alternate or potentially multiple languages at once is unknown. We note that existing work has shown adapters and hypernetworks to be useful for multilingual adaptation \citep{pfeiffer-etal-2020-mad, ustun-etal-2020-udapter, ansell-etal-2021-mad-g}, suggesting that our approach may be effective for multilingual multitasking when combined with these existing multilingual adaptation methods. Finally, the nature of the hypernetwork approach means that it may be difficult to scale to massive multi-tasking on the scale of exT5 \citep{aribandi2022ext} or T0 \citep{sanh2022multitask}, as the hypernetwork itself has limited capacity to store tasks. We do not investigate how this approach scales with respect to tasks, instead focussing on how the hypernetwork improves positive transfer and mitigates negative transfer for tasks whose positive and negative transfer effects are well-known.

\bibliography{anthology,custom}
\bibliographystyle{acl_natbib}
\appendix

\section{Baseline Details}
\label{sec:expp_details}

\subsection{Single Adapter}

Our single adapter baseline involves placing parallel adapters \citep{he2022towards} in each layer of the encoder and decoder and directly learning the parameters. For most tasks, we make use of adapters with size 410 in order to keep the trainable parameter budget equivalent to our approach. For MRQA, we achieve better results with adapters of size 800 in the encoder and size 36 in the decoder.

\subsection{Task Hypernet}

This model uses two hypernetworks to produce adapters for the encoder and decoder respectively, based on a learnt task embedding. The hypernetwork layout and adapter placement are identical to the one proposed in our method, with the exception that the initial input is a task embedding and not the mean-pooled encoder output. The inclusion of this baseline shows that the benefits of our method come from the use of input-specific decoders instead of improved adapter placement. For this approach, we use adapters of size 50 and hypernetworks of size 100, keeping the overall number of trainable parameters close to our approach.

\subsection{Modular Task Hypernetwork}

This is the method proposed by \citet{ponti2022combining}, which learns a mapping from task id to a sparse set of skills, each of which corresponds to a set of LoRA \citep{hu2022lora} parameters. These parameters are then combined to create task-specific adaptations. We set the number of skills $|\mathcal{S}| = 4$ and use a two-speed learning rate as suggested by the authors. We set the main learning rate as $3e-4$, and the secondary learning rate for $Z$ as $1e-2$. We use the same learning rate schedule as detailed in section \ref{sec:exp_details}. The rank of the LoRA adaptations is set to 16, following the defaults used in \citet{ponti2022combining}.

\section{Training Details}
\label{sec:training_details}
We run experiments on a single NVIDIA A100 80GB GPU, with all reported results from one training run using the provided hyperparameters. A single training run (including preprocessing and all evaluation) of our approach on the GLUE benchmark (with T5$_\text{large}$ v1.1 + LM as the underlying model) takes 18 hours. A single training run of our approach on the MRQA datasets (with T5$_\text{base}$ v1.1 + LM as the underlying model), including final evaluation, takes 13 hours. A training single run of our model on the Summarisation and NLI datasets (with T5$_\text{base}$ v1.1 + LM as the underlying model), including evaluation, takes 22.5 hours. We note that T5$_\text{base}$ v1.0 (or `vanilla') has roughly 220 million parameters total, T5$_\text{base}$ v1.1 + LM has roughly 250 million parameters total, and T5$_\text{large}$ v1.1 + LM has roughly 800 million parameters total.

\section{Encoder-Only Model Details}
\label{sec:encoder_only_expl}
The encoder-only models used in Table \ref{tab:enc_only_glue} consist of a T5 encoder with adapters inserted as done for the adapter-only and our model. We train a unique linear layer per task that maps from the hypernetwork embedding to logits, which are passed through a softmax layer to make a final prediction. We initialise the T5 encoder with the trained adapter layers in encoders, using the checkpoints reported in Table \ref{tab:glue_11}. This encoder is kept frozen during training. To match the MLP used to generate the hypernetwork embeddings in equation \ref{embed}, we place an identical MLP on top of the adapter-augmented T5 encoder and pass its output to the linear classifier. This MLP is trained along with the classifier. To train, we use the AdamW optimiser with a learning rate of 2e-5 for 3 epochs with linear learning rate warmup and decay with 500 warmup steps.

\section{Parameters}

We calculate the number of parameters for each method as follows. In each case, $l$ is the number of layers, $d$ the hidden size of the model, $a$ the adapter dimension, $t$ the number of tasks, and $b$ the hypernetwork bottleneck size. For simplicity, we will assume that the encoder and decoder adapter sizes are the same below, but it is straightforward to calculate the parameters used in the encoder and decoder separately and sum these for cases where the encoder and decoder sizes differ. We leave out bias parameters for simplicity.

\subsection{Adapters}

We only consider placing adapters in the feedforward module of the transformer layer. Every layer has one adapter consisting of two linear layers and a non-linearity, giving $l(2ad + a + d)$ overall new parameters.

\subsection{Task Hypernetwork}

For the task-embedding-based hypernetwork approach, we generate all adapters with two hypernetworks, one for the encoder and one for the decoder. the main parameter costs come in the form of the final layer of the hypernetwork, which consists of four linear layers producing the weights and biases of the two linear layers making up the adapter, thus costing $b(2ad + a + d)$. Given task embedding size $e_t$ and layer embedding size $e_l$, the total cost of the hypernetwork is $te_t + le_l + (e_t + e_l)b + b(2ad + a + d)$. This is then multiplied by two as we have hypernetworks for the encoder and decoder.

\subsection{Encoder-conditioned Decoders}

The cost of the decoder is similar to the task hypernetwork, except we add the cost of the MLP and use the hidden size of the model instead of the task embedding size: $2d^2 + (d + e_l)b + b(2ad + a + d)$. Note $b(2ad + a + d) \gg d^2$ for the adapter and model size choices used in our work. The encoder costs the same as the adapters case: $l(2ad + a + d)$. Our overall cost is simply the sum of the two.

\section{Dataset Statistics}

In Tables \ref{tab:glue_data_split}, \ref{tab:mrqa_data_split} and \ref{tab:xsum_data_splits} provide some summary statistics of each dataset used and split sizes.

\begin{table}
\centering
\begin{adjustbox}{width=.5\textwidth}
\begin{tabular}{l|ccc}
\hline
Dataset & Train Split Size & Validation Split Size & Test Split Size \\ \hline
CoLA & 8551 & 521 & 522 \\
SST-2 & 66349 & 1000 & 872 \\
STS-B & 5749 & 750 & 750 \\
MRPC & 3668 & 204 & 204 \\
QQP & 362846 & 1000 & 40430 \\
MNLI & 392702 & 9832 & 9815 \\
QNLI & 103743 & 1000 & 5463 \\
RTE & 2490 & 138 & 139 \\
\hline
\end{tabular}
\end{adjustbox}
\caption{Summary statistics of splits used when evaluating GLUE.}\label{tab:glue_data_split}
\end{table}
\begin{table}
\centering
\begin{adjustbox}{width=.5\textwidth}
\begin{tabular}{l|ccc}
\hline
Dataset & Train Split Size & Validation Split Size \\ \hline
SQuAD & 86588 & 10507 \\
HotpotQA & 72928 & 5901 \\
TriviaQA & 61688 & 7785 \\
NewsQA & 74160 & 4212 \\
SearchQA & 117384 & 16980 \\
Natural Qs & 104071 & 12836 \\
BioASQ & - & 1504 \\
DROP & - & 1503 \\
DuoRC & - & 1501 \\
RACE & - & 674 \\
Relation Ext. & - & 2948 \\
TextbookQA & - & 1503 \\
\hline
\end{tabular}
\end{adjustbox}
\caption{Summary statistics of splits used when evaluating MRQA, before applying preprocessing.}\label{tab:mrqa_data_split}
\end{table}
\begin{table}
\centering
\begin{adjustbox}{width=.5\textwidth}
\begin{tabular}{l|ccc}
\hline
Dataset & Train Split Size & Validation Split Size & Test Split Size \\ \hline
XSum & 204045 & 11332 & 11334 \\
CNN/Daily Mail & 287113 & 13368 & 11490 \\
Wiki Lingua & 99020 & 13823 & 28614 \\
MNLI & 392702 & 9832 & 9815 \\
Abductive NLI & 168654 & 1000 & 1532 \\
Adversarial NLI & 162865 & 3200 & 3200 \\
\hline
\end{tabular}
\end{adjustbox}
\caption{Summary statistics of splits used when evaluating summarisation and NLI datasets. Note we combine all Adversarial NLI rounds (r1, r2, r3).}\label{tab:xsum_data_splits}
\end{table}

\section{Full GLUE Ablation Results}
\label{sec:glue_ablation_full_results}

In Table \ref{tab:glue_full_ablations} we provide the full results from the ablations performed in Tables \ref{tab:glue_ablation} and \ref{tab:other_ablations}. Model names are in format <encoder type>-<decoder type>. `Generated' refers to encoder-conditioned adapters, `manual' regular adapters, and `task' task-conditioned adapters.

\begin{table*}
\centering
\begin{adjustbox}{width=1\textwidth,center=\textwidth}
\begin{tabular}{l|c|llllllll|l}
\hline
Model & \begin{tabular}{@{}c@{}}\% Trainable \\ Param.\end{tabular} & CoLA & SST-2 & STS-B & MRPC & QQP & MNLI & QNLI & RTE & Avg\\\hline
Full Finetuning & 100\% & \textbf{63.6} & 94.8 & 92.0 / 91.6 & 88.7 / 91.8 & \textbf{92.2 / 89.5} & 88.6 & 93.3 & 77.5 & 86.3 \\
Manual-Manual & 2.9\% & 58.5 & 95.3 & 91.7 / 91.4 & \textbf{89.7} / 92.5 & 91.2 / 88.3 & 89.8 & 94.0 & 81.2 & 86.4 \\
Manual-Task & 1.7\% & 54.6 & 96.0 & 91.9 / 91.9 & \textbf{89.7} / \textbf{92.6} & 91.0 / 88.1 & \textbf{90.0} & 94.1 & \textbf{83.3} & 86.3 \\
Manual-Generated & 2.9\% & 58.5 & \textbf{96.6} & 91.8 / 91.7 & \textbf{89.7} / 92.5 & 91.3 / \textbf{89.5} & 89.8 & \textbf{94.5} & 82.6 & \textbf{86.9} \\
Task-Manual & 1.7\% & 8.2 & 80.7 & 38.4 / 38.4 & 69.0 / 80.1 & 78.5 / 74.2 & 61.9 & 69.8 & 51.5 & 57.7 \\
Task-Generated & 3.0\% & 0.0 & 83.1 & 38.4 / 42.5 & 69.5 / 81.0 & 90.6 / 87.6 & 63.0 & 60.1 & 52.2 & 57.9 \\
Task-Task & 2.7\% & 0.0 & 82.1 & 16.4 / 16.4 & 70.4 / 81.4 & 89.8 / 86.5 & 56.6 & 64.8 & 50.7 & 54.3 \\
Generated-Manual & 3.3\% & 50.3 & 95.7 & 90.6 / 90.8 & 82.8 / 87.6 & 89.8 / 86.5 & 89.1 & 92.2 & 76.1 & 83.4 \\
Generated-Task & 3.3\% & 41.6 & 95.3 & 82.4 / 92.7 & 83.1 / 88.3 & 89.5 / 86.2 & 88.6 & 92.1 & 79.7 & 82.3 \\
Generated-Generated & 4.4\% & 45.4 & 95.8 & 88.6 / 88.4 & 84.2 / 88.5 & 89.7 / 86.3 & 89.5 & 91.8 & 76.8 & 82.8 \\ \hline
Manual-Generated, no MLP & 2.6\% & 62.0 & 96.0 & \textbf{92.2 / 92.3} & 86.7 / 90.3 & 91.1 / 88.2 & \textbf{90.0} & 93.4 & 81.9 & 86.7 \\
\begin{tabular}{@{}l@{}}Manual-Generated,\\ post-layernorm adapter input\end{tabular}
 & 2.9\% & 45.0 & 94.8 & 90.7 / 91.1 & 82.3 / 87.5 & 89.8 / 86.7 & 89.3 & 92.8 & 79.7 & 83.2 \\
\hline
\end{tabular}
\end{adjustbox}
\caption{Performance of models using T5$_{\text{large}}$ v1.1 + LM as the base model on GLUE test set splits described in section \ref{sec:data_glue}.  Trainable parameters is \% of the trainable parameters used compared to fully finetuning T5.}\label{tab:glue_full_ablations}
\end{table*}

\section{MRQA Varied Encoder Results}
\label{sec:mrqa_full_results}

In Tables \ref{tab:mrqa_in_full_encoder} and \ref{tab:out_mrqa_full_encoder} we provide results from experiments varying encoder and decoder sizes, as mentioned in section \ref{sec:mrqa_Res}. For the final results reported in Tables \ref{tab:in_mrqa} and \ref{tab:out_mrqa}, we use an encoder size of 512 for adapters and the hyperdecoder.

\begin{table*}[]
\begin{adjustbox}{width=1\textwidth,center=\textwidth}
\begin{tabular}{lcccccccccc|c}
\hline
Model &\begin{tabular}{@{}c@{}}Encoder \\ Adapter \\ Size\end{tabular} & \begin{tabular}{@{}c@{}}Decoder \\ Adapter \\ Size\end{tabular} & \begin{tabular}{@{}c@{}}Hypernet \\ Bottleneck \\ Size\end{tabular} & \begin{tabular}{@{}c@{}}\% Trainable \\ Param.\end{tabular} & SQuAD & HotpotQA & TriviaQA & NewsQA & SearchQA & Natural Qs & Avg \\ \hline
Hyperdecoder (ours) & 64 & 64 & 128 & 6.2\% & 91.1 & 78.9 & 73.5 & 67.7 & 82.6 & 77.6 & 78.6 \\
Hyperdecoder (ours) & 512 & 36 & 72 & 5.9\% & \textbf{91.3} & \textbf{79.9} & 75.0 & \textbf{68.6} & \textbf{82.9} & \textbf{79.1} & \textbf{79.5} \\
Hyperdecoder (ours) & 800 & 2 & 16 & 6.2\% & 90.8 & 79.5 & 75.0 & \textbf{68.6} & \textbf{82.9} & 78.6 & 79.2 \\
Adapter & 370 & 370 & - & 5.5\% & 88.7 & 75.9 & 67.4 & 64.9 & 79.5 & 74.3 & 75.1 \\
Adapter & 512 & 225 & - & 5.5\% & 90.8 & 79.5 & 75.0 & \textbf{68.6} & \textbf{82.9} & 78.6 & 79.2 \\
Adapter & 800 & 2 & - & 6.0\% & 91.0 & 79.6 & \textbf{75.5} & \textbf{68.6}  & \textbf{82.9} & 78.9 & 79.4 \\\hline
\end{tabular}
\end{adjustbox}
\caption{F1 score of models on in-domain MRQA validation split. All models use the T5$_{\text{base}}$ v1.1 + LM checkpoint as a starting point.}\label{tab:mrqa_in_full_encoder}
\end{table*}

\begin{table*}[]
\begin{adjustbox}{width=1\textwidth,center=\textwidth}
\begin{tabular}{lcccccccccc|c}
\hline
Model &\begin{tabular}{@{}c@{}}Encoder \\ Adapter \\ Size\end{tabular} & \begin{tabular}{@{}c@{}}Decoder \\ Adapter \\ Size\end{tabular} & \begin{tabular}{@{}c@{}}Hypernet \\ Bottleneck \\ Size\end{tabular} & \begin{tabular}{@{}c@{}}\% Trainable \\ Param.\end{tabular} & BioASQ & DROP & DuoRC & RACE & Relation Ext. & TextbookQA & Avg \\ \hline
Hyperdecoder (ours) & 64 & 64 & 128 & 6.2\% & 66.2 & 43.3 & 57.8 & 46.8 & 85.4 & 58.4 & 59.7 \\
Hyperdecoder (ours) & 512 & 36 & 72 & 5.9\% & 65.8 & 47.0 & \textbf{58.1} & 46.3 & 84.3 & \textbf{59.5} & \textbf{60.2} \\
Hyperdecoder (ours) & 800 & 2 & 16 & 6.2\% & 65.7 & 44.1 & 55.2 & 45.4 & \textbf{84.6} & 54.9 & 58.3 \\
Adapter & 370 & 370 & - & 5.5\% & 63.4 & 33.3 & 56.4 & 44.8 & 83.1 & 56.9 & 56.3 \\
Adapter & 512 & 225 & - & 5.5\% & 66.0 & \textbf{48.2} & 55.6 & \textbf{47.7} & 84.0 & 57.5 & 59.8 \\
Adapter & 800 & 2 & - & 6.0\% & \textbf{66.4} & 43.7 & 55.9 & 47.2 & 84.5 & 55.6 & 58.9 \\\hline
\end{tabular}
\end{adjustbox}
\caption{F1 score of models on out-of-domain MRQA validation split. All models use the T5$_{\text{base}}$ v1.1 + LM checkpoint as a starting point.}\label{tab:out_mrqa_full_encoder}
\end{table*}

\section{Full Summarisation and NLI Results}
\label{sec:xsum_nli_full_results}
In Table \ref{tab:xsum_nli_all_results} we provide a breakdown of the results reported in Table \ref{tab:xsum_nli}.

\begin{table*}[]
\begin{adjustbox}{width=\textwidth}
\centering
\begin{tabular}{lcccccccccc}
\hline
Model & \begin{tabular}{@{}c@{}}\% Trainable \\ Param.\end{tabular} & XSUM R2 & CNN/Daily Mail R2 & Wiki Lingua R2 & MNLI & Ab. NLI & Ad. NLI & Sum. Avg. & NLI Avg. & Overall Avg. \\ \hline
Full-finetuning & 100\% & \textbf{17.4} & \textbf{19.3} & \textbf{18.4} & 85.6 & 64.3 & 44.3 & \textbf{18.4} & 64.8 & 41.6 \\
Adapter & 5.5\% & 15.4 & 18.7 & 16.7 & 86.2 & 65.2 & 45.0 & 16.9 & 65.5 & 41.2 \\
Adapter (enc-heavy) & 5.5\% & 15.5 & 18.7 & 16.8 & \textbf{86.5} & \textbf{67.2} & \textbf{45.2} & 17.0 & \textbf{66.3} & \textbf{41.7} \\
Hyperdecoder (ours) & 6.2\% & 14.6 & 18.2 & 16.0 & 85.5 & 63.8 & 44.1 & 16.3 & 64.5 & 40.4 \\
Hyperdecoder (ours) (enc-heavy) & 5.9\% & 15.3 & 18.6 & 16.5 & 86.3 & 66.4 & 44.1 & 16.8 & 65.6 & 41.2 \\
Task Hypernet & 10.4\% & 11.6 & 17.0 & 13.4 & 83.9 & 61.5 & 40.5 & 14.0 & 62.0 & 38.0 \\
\hline
\end{tabular}
\end{adjustbox}
\caption{Performance across Summarisation and NLI datasets using T5$_{\text{base}}$ v1.1 + LM. Summarisation scores are Rouge2 scores, while NLI scores are accuracy using test splits described in section \ref{sec:data_xsum_nli}. `Enc-heavy' variants use encoder adapter sizes of 512 and the matching decoder/hypernetwork sizes in Table \ref{tab:mrqa_in_full_encoder}.} \label{tab:xsum_nli_all_results}
\end{table*}

\nocite{warstadt2018neural, socher2013recursive, dolan2005automatically, agirre2007semantic, williams2018broad, rajpurkar2016squad, dagan2006pascal, bar2006second, giampiccolo2007third, bentivogli2009fifth, levesque2011winograd}
\nocite{yang-etal-2018-hotpotqa, kwiatkowski-etal-2019-natural, trischler-etal-2017-newsqa, rajpurkar2016squad, searchqa, joshi-etal-2017-triviaqa, bioasq, dua-etal-2019-drop, saha-etal-2018-duorc, lai-etal-2017-race, levy-etal-2017-zero, Kembhavi_2017_CVPR}

\end{document}